\documentclass{article} 
\usepackage{iclr2026_conference,times}

\usepackage{amsmath,amsfonts,bm}

\def\eqref#1{equation~\ref{#1}}

\def\1{\bm{1}}

\DeclareMathAlphabet{\mathsfit}{\encodingdefault}{\sfdefault}{m}{sl}
\SetMathAlphabet{\mathsfit}{bold}{\encodingdefault}{\sfdefault}{bx}{n}

\usepackage{xcolor}
\usepackage{soul}
\usepackage{hyperref}
\usepackage{url}
\usepackage{algorithm}
\usepackage{algorithmic}

\usepackage{graphicx}
\usepackage[font=small,labelfont=bf]{caption}

\title{Smart-GRPO: Smartly Sampling Noise for Efficient RL of Flow-Matching Models}

\author{
Benjamin Yu\thanks{Equal contribution.} \\
Department of Computer Science \\
University of California, Los Angeles \\
Los Angeles, CA 90095, USA \\
\texttt{yubenjamin2022@ucla.edu}
\And
Jackie Liu\footnotemark[1] \\
Department of Computer Science \\
Brown University \\
Providence, RI 02912, USA \\
\texttt{ziyang\_liu@brown.edu}
\And
Justin Cui \\
Department of Computer Science \\
University of California, Los Angeles \\
Los Angeles, CA 90095, USA \\
\texttt{justincui@ucla.edu}
}

\iclrfinalcopy 
\begin{document}

\maketitle

\begin{abstract}
Recent advancements in flow-matching have enabled high-quality text-to-image generation. However, the deterministic nature of flow-matching models makes them poorly suited for reinforcement learning, a key tool for improving image quality and human alignment. Prior work has introduced stochasticity by perturbing latents with random noise, but such perturbations are inefficient and unstable. We propose Smart-GRPO, the first method to optimize noise perturbations for reinforcement learning in flow-matching models. Smart-GRPO employs an iterative search strategy that decodes candidate perturbations, evaluates them with a reward function, and refines the noise distribution toward higher-reward regions. Experiments demonstrate that Smart-GRPO improves both reward optimization and visual quality compared to baseline methods. Our results suggest a practical path toward reinforcement learning in flow-matching frameworks, bridging the gap between efficient training and human-aligned generation.

\end{abstract}

\section{Introduction}

Flow-matching models \citep{lipman2023flow} have recently been introduced as an alternative to diffusion-based generative models \citep{ho2020denoising}, offering more stable training and deterministic sampling. While large-scale pre-training enables these models to produce high-quality outputs, it is often insufficient for ensuring alignment with human preferences. Reinforcement learning with human feedback (RLHF) \citep{ouyang2022training}, originally developed for aligning large language models, has since been adapted to generative vision models, including diffusion architectures \citep{black2023training, yang2024using}.

Extending RL to flow-matching models, however, presents distinct challenges. The deterministic nature of flow-matching sampling is fundamentally misaligned with the stochasticity required for policy optimization. Flow-GRPO \citep{liu2025flow} addresses this by introducing random perturbations to the inputs prior to denoising, thereby enabling the use of Group Relative Policy Optimization (GRPO) \citep{shao2024deepseekmath}. While this modification permits reinforcement learning, it is intrinsically inefficient: most noise seeds sampled uniformly at random produce low-reward generations that contribute little to policy improvement, resulting in wasted training signal. This observation motivates a more principled treatment of noise selection as a key factor in improving the efficiency of RL for flow-matching models.

Prior works have focused primarily on improving training stability \citep{wang2025pref, xue2025dancegrpo}, efficiency through optimization strategies \citep{li2025mixgrpo, li2025branchgrpo}, or generative fidelity \citep{he2025tempflow}. Our work specifically focuses on the noise used to perturb the inputs. We hypothesize that, by directly optimizing the sampling of input noise, we provide a complementary pathway for improving both efficiency and alignment in reinforcement learning for flow-based generative models.

In this work, we introduce \textbf{Smart-GRPO}, a framework that augments Flow-GRPO with reward-guided noise selection. Our central hypothesis is that noise seeds vary in their contribution to effective learning, and that preferentially sampling informative seeds can accelerate convergence. Smart-GRPO employs a pretrained reward model to evaluate candidate noise seeds and selects those predicted to yield higher-quality generations. This procedure can be viewed as constructing an adaptive curriculum over the noise distribution, where training gradually emphasizes seeds that produce more informative trajectories. By iteratively refining the noise distribution in this manner, Smart-GRPO improves the efficiency of policy optimization while maintaining compatibility with existing RLHF pipelines.  

\begin{figure}[h]
    \centering
    \includegraphics[width=1\linewidth]{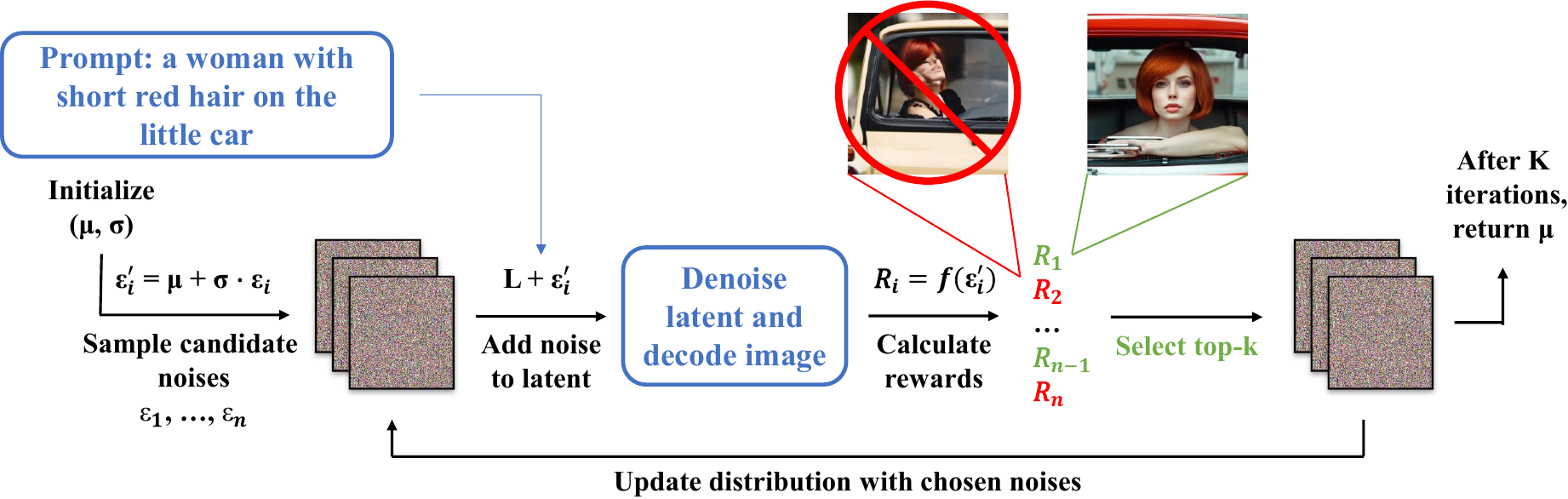}
    \caption{Overview of Smart-GRPO. The method begins by initializing a Gaussian noise distribution parameterized by $(\mu, \sigma)$. At each iteration, candidate noise samples are drawn, applied to perturb the latent representation, denoised for one step, and decoded into images. A reward model evaluates the resulting images, and the top-k
k noise samples are used to update the distribution parameters. After K iterations, the final mean 
$\mu$ is selected as the optimized noise for GRPO training.}
    \label{fig:smart-grpo}
\end{figure}

\section{Related Works}

\textbf{Flow-matching models}: Let $x_0 \in X_0$ be a sample from a true distribution and let $x_1 \in X_1$ be a sample from a known distribution (e.g. a Gaussian). Flow-matching models \citep{esser2024scaling} define a path between the data and the noise as a linear interpolation:

\begin{equation}
\label{interpolation}
x_t = (1-t)x_0 + t x_1, \quad t \in [0,1].
\end{equation}

Taking the derivative with respect to  t yields the target velocity field:

\begin{equation}
\frac{dx_t}{dt} = x_1 - x_0.
\end{equation}

The goal is then to learn a parameterized velocity predictor $v_{\theta}(x_t,t)$ that approximates this ground-truth field. This is achieved by minimizing the following loss \citep{lipman2023flow}:

\begin{equation}
\mathbb{L}(\theta) = \mathbb{E}_{t, x_0, x_1}\big[ || (x_1 - x_0) - v_\theta(x_t, t) ||^2 \big].
\end{equation}

Compared to diffusion \citep{ho2020denoising} models, which learn a score function or directly predict noise, flow-matching instead learns the velocity of the probability flow ODE. This provides a more direct parameterization of the generative process, and in practice can lead to faster and more stable training.

\textbf{Reinforcement Learning for Generative Models}:
Reinforcement learning from human feedback (RLHF) \citep{ouyang2022training} has become a standard approach for aligning large language models with human preferences. The framework typically involves first training a reward model on collected preference data, and then fine-tuning the language model using Proximal Policy Optimization (PPO) \citep{schulman2017proximal}. More recently, alternatives such as Direct Preference Optimization (DPO) \citep{rafailov2023direct} and Group Relative Policy Optimization (GRPO) \citep{shao2024deepseekmath} have been proposed, offering simpler and more flexible formulations of preference-based training.

Recently, reinforcement learning has also been adapted to diffusion models, which present unique challenges due to their iterative denoising process. Approaches such as Denoising Diffusion Policy Optimization (DDPO) \citep{wallace2024diffusion} and Direct Preference for Denoising Diffusion Policy Optimization (D3PO) \citep{yang2024using} extend preference-based optimization to the diffusion setting, enabling alignment with human or other task-specific objectives.
 
\textbf{Reinforcement Learning for Flow-matching models}: Due to the deterministic nature of flow-matching models, they are not intrinsically designed for reinforcement learning. The probability flow ODE deterministically maps inputs to outputs, leaving little room for the stochastic exploration that reinforcement learning requires. This mismatch makes direct application of standard policy optimization methods ineffective.

Flow-GRPO \citep{liu2025flow} addresses this by converting the deterministic probability flow ODE into an equivalent stochastic differential equation (ODE-to-SDE), which injects randomness while preserving the model’s marginal distributions, and by introducing a denoising reduction strategy that reduces the number of denoising steps during training while keeping the full schedule at inference. These modifications enable the incorporation of GRPO into flow-matching models. Empirically, Flow-GRPO achieves substantial gains in compositional image generation, text rendering, and human preference alignment, while maintaining image quality and minimizing reward hacking.

\section{Methods}

We introduce Smart-GRPO, an efficient algorithm for fine-tuning flow-matching models with reinforcement learning. Our method improves upon GRPO-style approaches by directly searching over the noise variables that determine the decoded output. Instead of perturbing latents with random noise (as in GRPO), Smart-GRPO searches for noise that maximizes reward in one-shot decoding. We treat the noise distribution as a parameterized search space. Instead of blindly perturbing latents, we iteratively refine a Gaussian noise distribution toward regions of higher reward using a Cross-Entropy Method (CEM)-like update.

\subsection{Algorithm}

Let $X_t$ denote the latent at timestep t, and let $f:X\to \mathbb{R}$ be a scalar reward function. Smart-GRPO proceeds as follows: We first initialize a Gaussian distribution over noise variables, parameterized by mean $\mu = 0$ and standard deviation $\sigma = I$. In each iteration, we sample K candidate noises as: 

\begin{equation}
    m_i = \mu + \sigma n_i, \space n_i \sim N(0, I)
\end{equation}

We then perturb the latent with the noise via the following equation:

\begin{equation}
    Z_i = X_t + \sqrt{-dt}\sigma_tm_i
\end{equation}

where $\sigma_t$ is the noise scale and $dt$ is the step size. To evaluate the effect of each perturbation, we form a one-step approximation of the decoded image using the predicted velocity $v_{\theta}$. 

\begin{equation}
    \label{approximation}
    x_0^{(i)} \approx z_i - tv_{\theta}(z_i, t)
\end{equation}

 This is intended to provide a rough estimate of the final image without requiring the full reverse process. Note that at earlier timesteps (high noise levels), the one-step approximation produces near-random outputs, making reward evaluation unreliable. Smart-GRPO is therefore most effective at later timesteps where the latent has a stronger correlation with the decoded image
 
 We then decode the image, and calculate the reward $R_i = f(x_0^{(i)})$. We then select the top $T = \lfloor P \cdot K \rfloor$, where $P \in [0, 1]$ candidates with the highest rewards, using these noises to update the $\mu$ and $\sigma$ used to sample. This process is repeated $N$ times.
 
This update step shifts the distribution toward higher-reward regions while adaptively controlling its spread, ensuring a balance between exploration and exploitation.

Once this process is complete, either the mean noise $\mu$ or a final sample $m = \mu + \sigma n$ is drawn to be used to perturb the latent for training. 

\begin{algorithm}[H]
\caption{Smart-GRPO}\label{alg:smart-grpo}
\begin{algorithmic}[1]
\REQUIRE Latent image $X_t$, number of sampled noises $K$, number of iterations $N$, saving fraction $P \in [0,1]$, reward function $f(z): X \to \mathbb{R}$
\ENSURE Optimized latent mean $\mu$ or sampled latent $m = \mu + \sigma \cdot n$

\STATE Initialize $\mu = 0$, $\sigma = I$ of the shape of latent variable

\FOR{$n = 1$ \textbf{to} $N$}
    \STATE Sample $K$ random noises $\{n_i\}_{i=1}^{K}$ and compute modified noises $m_i = \mu + \sigma \cdot n_i$
    \STATE Perturb the latent with noise:
        \[
        {Z}_i = X + \sqrt{-dt} \cdot \sigma_t \cdot m_i
        \]
    \STATE Decode latents $Z_i$ from $m_i$ and compute reward $R_i = f(Z_i)$
    
    \STATE Select top $T = \lfloor P \cdot K \rfloor$ noises with highest rewards
    \STATE Update mean and standard deviation:
    \[
        \mu = \frac{1}{T} \sum_{i=1}^{T} m_i, \quad
        \sigma^2 = \frac{1}{T} \sum_{i=1}^{T} (m_i - \mu)^2
    \]
\ENDFOR

\RETURN $\,\mu \,$ or $\, m = \mu + \sigma n$
\end{algorithmic}
\end{algorithm}
Unlike prior methods that inject noise uniformly at random, Smart-GRPO requires no architectural changes and introduces only a lightweight noise-selection loop, making it straightforward to integrate into existing RLHF pipelines.
\section{Experiments}

This section describes the methods used to empirically evaluate whether Smart-GRPO improves performance of flow-matching models. To show this, we utilize two baselines and train our model on two reward functions and analyze results.

\textbf{Baselines}: To compare the performance of our algorithm on fine-tuning flow-matching models, we have two baselines: base Stable-Diffusion 3.5-M \citep{esser2024scaling}, base Stable-Diffusion 3.5-L \citep{esser2024scaling}, base FLUX.1-dev \citep{batifol2025flux},  and Stable-Diffusion 3.5-M fine-tuned with Flow-GRPO \citep{liu2025flow} without our algorithm. 

We selected ImageReward and Aesthetic Score as reward functions because they capture complementary aspects of text-to-image generation. ImageReward is a general-purpose model trained to evaluate prompt-image alignment, visual fidelity, and harmlessness, making it a broad measure of generation quality. In contrast, the Aesthetic Score directly targets visual appeal, reflecting how pleasing an image is to human perception. Using both rewards allows us to evaluate Smart-GRPO across semantic alignment and visual quality, demonstrating its effectiveness under different alignment objectives.  

We choose a prompt dataset of 3000 prompts sampled from datasets provided by Flow-GRPO, generated from GenEval scripts \cite{ghosh2023geneval} to train our models on, which was randomly sampled. We split our dataset into a training and evaluation dataset of 2700 training prompts and 300 evaluation prompts. See \ref{hyperparameters} for more details regarding specific hyperparameters used during training. 

\subsection{Analysis}

\begin{figure}[h]
    \centering
    \includegraphics[width=1\linewidth]{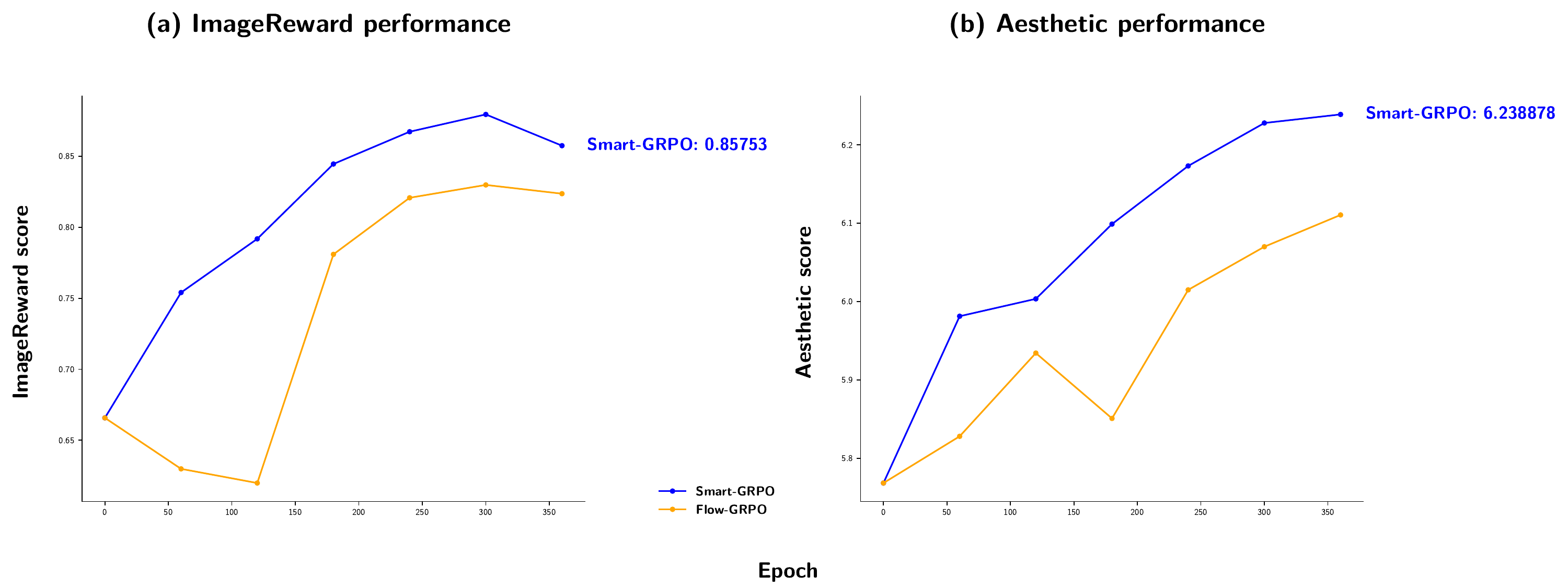}
    \caption{Training results of Smart-GRPO over 360 epochs. Figure \textbf{(a)} is trained with ImageReward, and Figure \textbf{(b)} is trained using the Aesthetic score}
    \label{fig:training-results}
\end{figure}

For both rewards we experimented on, our method has both better performance and more stable compared to base Flow-GRPO. Figure~\ref{fig:training-results} shows that Smart-GRPO consistently improves ImageReward scores across training epochs, converging faster and achieving higher final reward than Flow-GRPO. Over our evaluation dataset, Smart-GRPO consistently outperforms the baseline models, as shown in Table~\ref{table:results-table}

\begin{table}[t]
\begin{center}
\begin{tabular}{lll}
\\ \hline \\
\multicolumn{1}{c}{\bf Model}  &\multicolumn{1}{c}{\bf ImageReward} &\multicolumn{1}{c}{\bf Aesthetic Score}
\\ \hline \\
Stable Diffusion 3.5M         &0.6658 &5.769 \\
Stable Diffusion 3.5L             &-0.0310&5.602\\
FLUX.1-dev             &0.5121  &6.093 \\

SD 3.5M (with Flow-GRPO) &\sethlcolor{green!50}\hl{0.8237} &\sethlcolor{green!50}\hl{6.111} \\
SD 3.5M (with Smart-GRPO) &\sethlcolor{blue!50}\hl{0.8575} &\sethlcolor{blue!50}\hl{6.238} \\

\\ \hline \\

\end{tabular}

\caption{Smart-GRPO model results over evaluation dataset. Dataset consists of 1000 sample prompts generated from GenEval scripts, provided in Flow-GRPO's repository. Results are means over scores of generated images from prompt dataset. Best results are highlighted in \sethlcolor{blue!50}\hl{blue}, second best highlighted in \sethlcolor{green!50}\hl{green}. }
\label{table:results-table}
\end{center}
\end{table}

\subsection{Ablation Study}

\begin{figure}[h]
    \centering
    \includegraphics[width=1\linewidth]{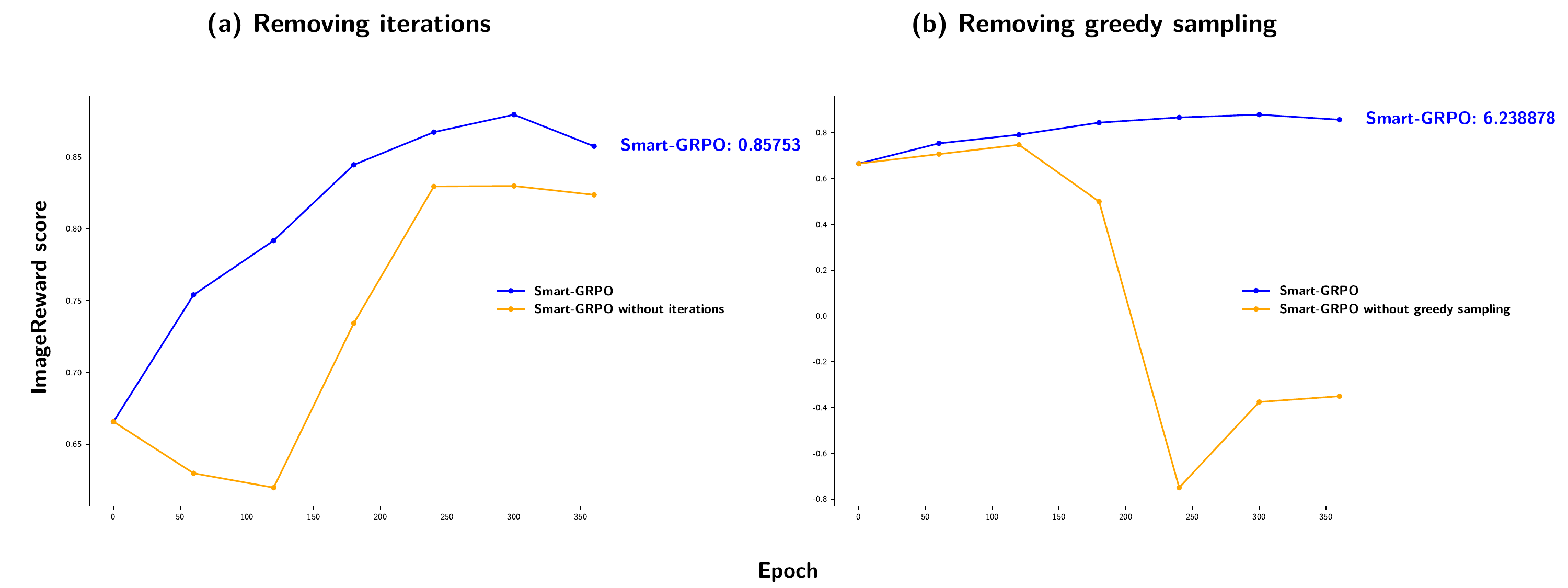}
    \caption{Figure for ablation studies}
    \label{fig:ablation}
\end{figure}

To better understand the contribution of Smart-GRPO’s core mechanisms, we conduct ablation studies on its two central components: \textbf{iterative refinement} and \textbf{greedy noise selection}. These experiments are designed to address two key questions: (1) Does iterative refinement across multiple rounds of sampling improve performance beyond a one-shot aggregation of noise samples? (2) Is greedy selection of high-reward noise essential, or can random updates achieve comparable results? All ablations were performed under the same training hyperparameters and with the ImageReward reward function.  

\paragraph{Iterative refinement.}  
Instead of progressively updating the noise distribution, we evaluate a one-shot alternative in which 25 noise samples are drawn, the top 12 are selected, and their mean is used to perturb the latent. We compare this baseline to Smart-GRPO with $N=5$ iterations and $K=5$ sampled noises per iteration. As shown in Figure~\ref{fig:ablation}a, both approaches achieve similar performance, but Smart-GRPO consistently outperforms the one-shot baseline. This suggests that iterative refinement enables the model to repeatedly concentrate its sampling distribution around higher-quality noise regions, yielding stronger overall performance.  

\paragraph{Greedy noise selection.}  
To assess the role of greedy selection, we replace high-reward noise selection with random sampling, keeping $N=5$ and $K=5$. As shown in Figure~\ref{fig:ablation}b, this variant exhibits highly unstable training: repeatedly updating with low-quality noise often leads to collapse and degraded generations. In contrast, greedy selection stabilizes training by systematically steering updates toward high-reward regions.  

Together, these results validate Smart-GRPO’s design: iterative refinement progressively sharpens the sampling distribution, while greedy selection prevents detrimental drift and ensures stable convergence. More broadly, these findings highlight the importance of treating noise as an optimization variable, reinforcing Smart-GRPO’s central contribution and paving the way for future work on noise-aware reinforcement learning in generative models.  

\subsection{Sensitivity Analysis}

\begin{figure}[h]
    \centering
    \includegraphics[width=0.6\linewidth, height=0.2\textheight]{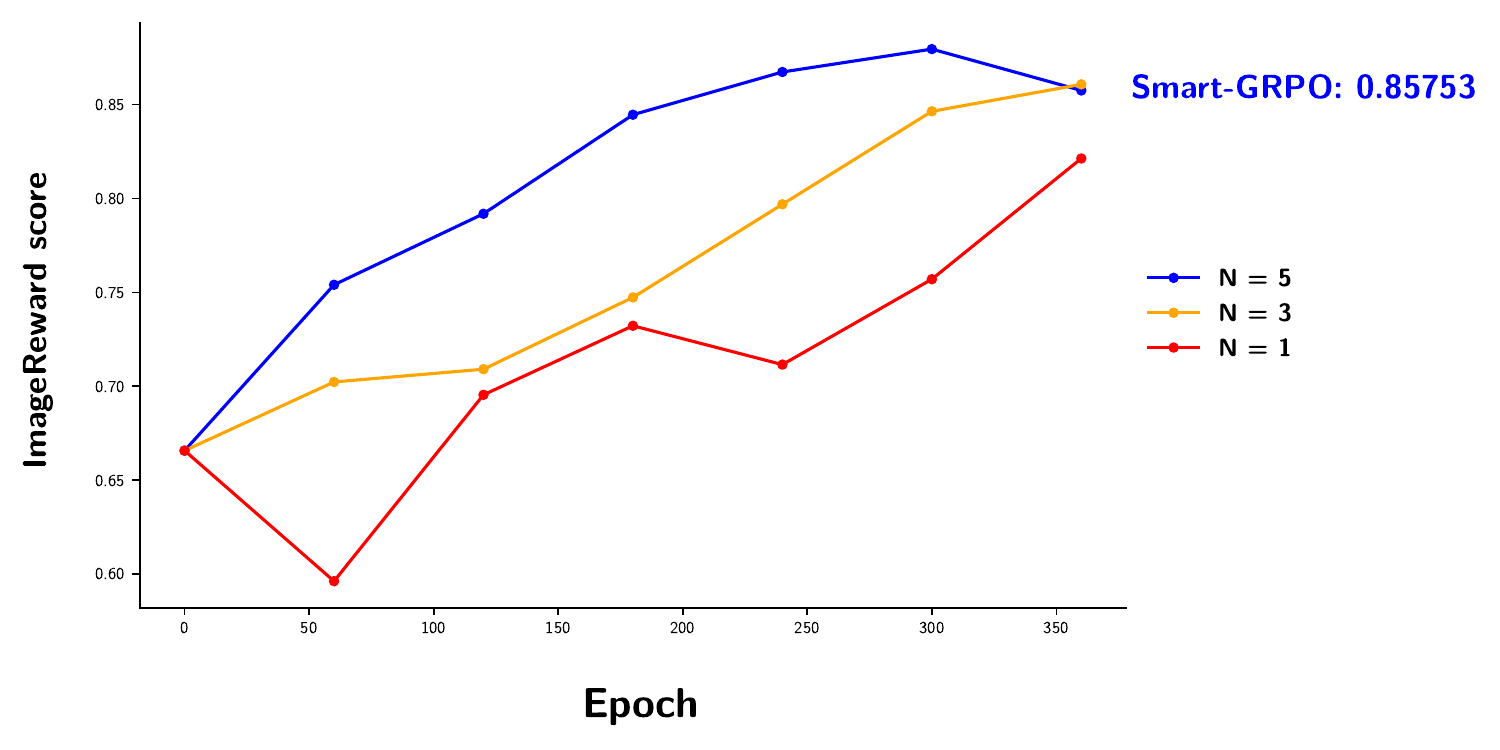}
    \caption{Sensitivity analysis for number of iterations used for Smart-GRPO. For 1 iteration, performance is unstable and fluctuates. When number of iterations increases, performance increases as iterations improve parameters more reliably.}
    \label{fig:sensitivity}
\end{figure}

We conducted a sensitivity analysis on the number of iterations to investigate how the choice of iterations influence training stability and reward performance.  

When only a single iteration ($N=1$) is used, Smart-GRPO effectively reduces to a one-shot update of the noise distribution. In this setting, the method provides small improvements over random perturbations, but the refinement process is too shallow: reward curves fluctuate noticeably across training, and final performance remains inconsistent.  

Increasing to three iterations ($N=3$) produces a marked change. Training becomes significantly more stable, with reward trajectories that are smoother and less noisy, and the models consistently achieve higher ImageReward and Aesthetic scores compared to $N=1$. This suggests that multiple rounds of refinement allow the noise distribution to more reliably concentrate probability mass in promising regions.  

With five iterations ($N=5$), Smart-GRPO reaches its strongest performance. Both ImageReward and Aesthetic scores converge to their highest values, and optimization proceeds in a stable and predictable manner. Here, the repeated refinement cycles appear to provide the algorithm with enough opportunities to progressively adjust the distribution toward high-reward samples without overfitting to noise.  

Taken together, these results highlight the importance of iterative refinement. By repeatedly resampling and updating, Smart-GRPO progressively guides the noise distribution toward high-quality solutions, improving both convergence speed and stability. While increasing $N$ beyond 5 may yield further gains, it also comes with higher computational cost. Our experiments suggest that $N \in \{3, 5\}$ strikes a practical balance between efficiency and performance. Due to computational constraints, we did not explore higher values of $N$, leaving a more extensive exploration of this trade-off to future work.  

\subsection{Limitations}

\begin{figure}[h]
    \centering
    \includegraphics[width=1\linewidth]{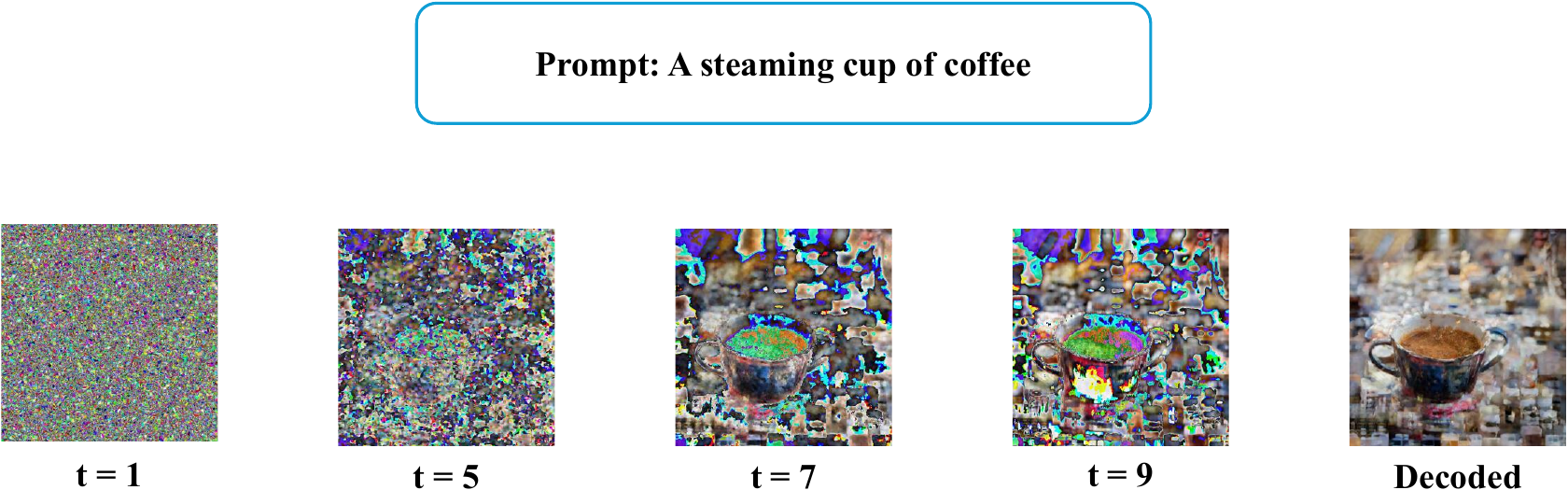}
    \caption{Intermediate approximations from Equation~\ref{approximation} during flow-matching generation of the prompt ‘A steaming cup of coffee’. Starting from a noise level of 0.6 and decoded over 10 steps, earlier timesteps yield outputs resembling noise, while later timesteps progressively form low-quality images. }
    \label{fig:timestep}
\end{figure}

While Smart-GRPO demonstrates promising improvements, it also has several limitations. First, the effectiveness of our approach depends heavily on the choice of reward function. Many reward models are not well calibrated to evaluate poor-quality or highly noisy images, which constrains their ability to guide the noise selection process. For example, our experiments with PickScore \citep{kirstain2023pick} and CLIPScore \citep{hessel2021clipscore} did not yield statistically significant gains, suggesting that these metrics may be ill-suited for reinforcement learning in high-noise regimes. A further limitation arises from the greedy approximation used in equation~\ref{approximation}: as illustrated in Figure~\ref{fig:timestep}, this approximation does not hold well at earlier timesteps. Two issues follow: (1) the reward model is not designed to reliably score low-quality images, and (2) the early approximations themselves often fail to capture a meaningful representation of the final image. Exploring alternative metrics to evaluate high-quality noise could potentially be beneficial in improving model generations. 

Second, due to computational constraints, we were unable to fully explore larger-scale experiments, longer training schedules, or higher values of  K and N, which could further clarify the method’s benefits.

\section{Conclusion}

We introduced Smart-GRPO, a simple yet effective framework for fine-tuning flow-matching generative models with reinforcement learning. By guiding noise sampling toward higher-reward regions, Smart-GRPO reduces wasted updates and improves both stability and convergence compared to existing methods. Importantly, this is the first approach to explicitly optimize the noise process for reinforcement learning in flow-matching models, offering a novel perspective on noise-aware training. We hope this work paves the way for future research on noise optimization, reward design, and scalable reinforcement learning for generative modeling.

\subsection*{Future Works}

For future work, we envision several directions. One is to design or identify reward functions that are more robust at distinguishing subtle improvements in image quality, especially in the presence of noise. Another is to investigate alternative strategies for noise selection beyond mean and variance updates, such as adaptive sampling or learned proposal distributions. Finally, scaling experiments to larger models and more diverse benchmarks would provide a clearer picture of the generality and practical impact of Smart-GRPO.

\section*{Impact Statement}

Smart-GRPO introduces a lightweight and efficient framework for reinforcement learning in flow-matching generative models. By directly optimizing the noise distribution, it reduces wasted training signal and achieves higher reward performance with fewer iterations. Because the method requires no architectural modifications and only a simple noise-selection loop, it can be seamlessly integrated into existing RLHF pipelines and deployed with modest computational resources.

In addition to these practical advantages, Smart-GRPO represents the first attempt to explicitly optimize the noise process for flow-matching models in the reinforcement learning setting. By reframing noise as an optimization variable, the method provides a novel perspective for improving generative modeling with reinforcement learning. This contribution opens a new line of research in noise-aware optimization, complementing advances in reward design and training objectives. Moreover, the generality of the approach suggests that similar techniques may be extended to diffusion models and other generative frameworks, paving the way for broader methodological innovations.

\bibliographystyle{iclr2026_conference}
\bibliography{iclr2026_conference}

\appendix
\section{Appendix}

\subsection{Hyperparameters}
\label{hyperparameters}

The hyperparameters used for training our models were largely copied from the hyperparameters used for Flow-GRPO. We only used 1 H100 GPU for training our model with GRPO. We initialized from the StabilityAI Stable Diffusion 3.5 Medium checkpoint.

All images were generated at a resolution of 512 × 512. During training, we used 10 sampling steps, while evaluation employed 40 sampling steps. We applied a classifier-free guidance scale of 4.5.

Each training batch contained 4 images per prompt, with an effective training batch size of 4. To ensure balanced gradient updates, we set the number of batches per epoch to 8, which yielded an even number of batches per epoch. Gradient accumulation was configured such that two updates occurred per epoch. The test batch size was fixed to 16.

We trained with 1 inner epoch per outer epoch, and sampled timesteps with a fraction of 0.99. Optimization included a KL loss term weighted by $\beta$ = 0.04. We enabled exponential moving average (EMA) of model weights.

We saved model checkpoints every 60 epochs and performed evaluation at the same frequency.

\end{document}